\documentclass[sn-mathphys,Numbered]{sn-jnl}

\usepackage{graphicx}%
\usepackage{multirow}%
\usepackage{amsmath,amssymb,amsfonts}%
\usepackage{amsthm}%
\usepackage{mathrsfs}%
\usepackage[title]{appendix}%
\usepackage{xcolor}%
\usepackage{textcomp}%
\usepackage{manyfoot}%
\usepackage{booktabs}%
\usepackage{algorithm}%
\usepackage{algorithmicx}%
\usepackage{algpseudocode}%
\usepackage{listings}%
\usepackage{longtable}
\usepackage{adjustbox}
\usepackage{array}

\theoremstyle{thmstyleone}%
\theoremstyle{thmstyletwo}%

\theoremstyle{thmstylethree}%

\begin{document}

\title{An empirical study of using radiology reports and images to improve ICU mortality prediction}


\author[1]{\fnm{Mingquan} \sur{Lin}}
\equalcont{These authors contributed equally to this work.}

\author[2]{\fnm{Song} \sur{Wang}}
\equalcont{These authors contributed equally to this work.}

\author[3]{\fnm{Ying} \sur{Ding}}

\author[4]{\fnm{Lihui} \sur{Zhao}}

\author[1]{\fnm{Fei} \sur{Wang}}

\author*[1]{\fnm{Yifan} \sur{Peng}}\email{yip4002@med.cornell.edu}

\affil[1]{\orgdiv{Department of Population Health Sciences}, \orgname{Weill Cornell Medicine}, \state{New York}, \country{USA}}

\affil[2]{\orgdiv{Cockrell School of Engineering}, \orgname{The University of Texas at Austin}, \state{Austin}, \country{USA}}

\affil[3]{\orgdiv{School of Information}, \orgname{The University of Texas at Austin}, \state{Austin}, \country{USA}}

\affil[4]{\orgdiv{Department of Preventive Medicine}, \orgname{Northwestern University}, \state{Chicago}, \country{USA}}


\abstract{\textbf{Background:} The predictive Intensive Care Unit (ICU) scoring system plays an important role in ICU management because it predicts important outcomes, especially mortality. Many scoring systems have been developed and used in the ICU. These scoring systems are primarily based on the structured clinical data in the electronic health record (EHR), which may suffer the loss of important clinical information in the narratives and images.

\textbf{Methods:} In this work, we build a deep learning based survival prediction model with multi-modality data to predict ICU mortality. Four sets of features are investigated: (1) physiological measurements of Simplified Acute Physiology Score (SAPS) II, (2) common thorax diseases pre-defined by radiologists, (3) BERT-based text representations, and (4) chest X-ray image features. We use the Medical Information Mart for Intensive Care IV (MIMIC-IV) dataset to evaluate the proposed model.

\textbf{Results:} Our model achieves the average C-index of 0.7829 (95\% confidence interval, 0.7620-0.8038), which substantially exceeds that of the baseline with SAPS-II features (0.7470 (0.7263-0.7676)). Ablation studies further demonstrate the contributions of pre-defined labels (2.00\%), text features (2.44\%), and image features (2.82\%).

\textbf{Conclusions:} Our model achieves a higher average C-index than the traditional machine learning methods under the same feature fusion setting, suggesting that the deep learning methods can outperform the traditional machine learning methods in ICU mortality prediction. These results highlight the potential of deep learning models with multimodal information to enhance ICU mortality prediction. We make our work publicly available at \url{https://github.com/bionlplab/mimic-icu-mortality}.}




\maketitle

\section{Introduction}

Predictive ICU scoring systems measure the disease severity to predict outcomes, typically mortality, of patients in the intensive care unit (ICU)~\cite{lipshutz2016predicting}. Such measurements help standardize research and compare the quality of patient care across ICUs. For example, the Acute Physiology and Chronic Health Evaluation~\cite{zimmerman2006acute}, Simplified Acute Physiology Score (SAPS) II~\cite{le1993new}, and Mortality Probability Model~\cite{teres1987validation} were created explicitly for the use in the ICU, validated in the critically ill, and relied primarily on the physiologic data to predict mortality. These scoring systems have been primarily based on structured clinical data including the risk factors used by the scoring system such as demongraphsics, vital signs and lab tests, which are frequently documented in the electronic health record (EHR).

The recent development of machine learning offers great potential to improve ICU mortality prediction~\cite{el2020intensive, ghassemi2015multivariate, zhao2020prediction, johnson2017real}. However, these studies used only structured, coded approaches for data entry, thus may result in the loss of significant clinical information typically contained in the narratives and images~\cite{murdoch2013inevitable, ching2018opportunities}. To overcome this issue, many studies focus on mining unstructured clinical notes for patient mortality prediction~\cite{mechcatie2018nursing, yang2021multimodal, grnarova2016neural}. However, most of these works were not compared with the current scoring system, making it challenging to compare these models fairly.

Moreover, the practice of modern medicine usually relies on multimodal information. Consequently, many feature fusion strategies were proposed to enhance the performance of prediction algorithms, such as early fusion, late fusion, and joint fusion~\cite{huang2020fusion}. Early fusion combines multimodal features into a single vector by concatenating or averaging~\cite{liu2017visual, liu2018prediction, liu2018bone}. Late fusion combines the predictions of multiple models to make the final decision~\cite{bakkali2020visual, reda2018deep, qiu2018fusion}. Joint fusion combines the features from the intermediate layer of the neural network with the features of other modalities. The loss during the training process will propagate back to the feature extraction neural network, thereby creating a better feature representation through training iterations~\cite{huang2020fusion, yala2019deep, kawahara2018seven, yoo2019deep}. Despite these encouraging findings, we note that most competitive approaches studied the classification tasks. Thus, the integration of text and images in the survival analysis framework remains an important yet, to date, insufficiently studied problem.

We, therefore, sought to overcome these limitations by (1) incorporating the potentials of natural language processing (NLP) and medical image analysis to identify the hidden features of critical illness among ICU patients in the radiology reports and chest X-rays that may not be found in the structured EHR fields~\cite{ford2016extracting}; and (2) investigating deep learning models that may provide the superior discrimination of ICU mortality predictions compared to traditional machine learning models~\cite{weissman2018inclusion}. Specifically, we first build the clinical prediction models to predict ICU mortality using the SAPS-II risk factors such as demographics, vital signs, and lab tests. These measurements were obtained in the first 24 hours of ICU admission. We then enrich the model with multimodal features extracted from radiology reports and chest X-rays. The radiology imaging and reading were studied in the first 24 hours. We hypothesize that including free texts and images provides better predictions of ICU mortality than including clinical measurements alone. Experiments on the MIMIC-IV dataset~\cite{johnsonalistairmimiciv} show that our multimodal models are substantially more accurate than the unimodal ones.

Our framework has several important strengths. First, we present a method to fuse multimodal data from EHR for ICU mortality prediction. Second, we demonstrate that our survival analysis model outperforms the existing clinical standards (SAPS-II). Third, we make our work publicly available for reproduction by others.

\section{Methods}

\subsection{Task}

We first formulate the survival analysis task, which predicts a patient's survival probability in ICU as a function of their features. We have $n$ patients $(x_{i},y_{i},\delta_{i})$. Each patient record consists of $d$ potential covariants $x_{i} \in R^{d}$, and the time $T_{i}$ when the death occurred or the time $C_{i}$ of censoring. Since death and censoring are mutually exclusive, we use the indicator $\delta_{i} \in \{0, 1\}$ and the observed survival time $y_{i}$, defined as below.

\begin{equation}\label{Eq: survivaltime}
y_i = min(T_i,C_i) = 
\begin{cases}
T_i& \text{if $\delta_i$ = 1}\\
C_i& \text{if $\delta_i$ = 0}
\end{cases}
\end{equation}

The goal is to estimate the survival probability $S_{i}(t) = Pr_{i}(T > t)$ of a patient who was not dead beyond time $t$.

In this study, we use one of the most popular survival analysis models, the Cox model~\cite{wang2017chestxray8}, where the survival function is assumed to be

\begin{equation}\label{Eq: survivalfunction}
S_i(t|x_i)=S_0(t)^{e^{\psi(x_i)}}.
\end{equation}

In this model, $S_{0}(t)$ is the baseline survival function that describes the risk for individuals with $x_{i} = \mathbf{0}$, and $\psi(x_{i}) = x_{i}\beta$ is the relative risk based on the covariants. Note that $S_{0}(t)$ is shared by all patients at time $t$. It is NOT associated with any individual covariants. The effect of the covariate values $x_{i}$ on the survival function is to raise it to a power given by the relative risk.

In the Cox model, $\psi(x_{i})$ has the form of a linear function, but we can also extend it to a non-linear risk function of a neural network, called the DeepSurv-based model. The DeepSurv-based model has three steps: features extraction, multimodal feature fusion, and survival analysis. The main difference between our model and the DeepSurv model in~\cite{katzman2018deepsurv} is that our deep network performs multimodal feature fusion. When there is only a single modality as input, our model is equivalent to the DeepSurv model. The detail of the neural network via feature fusion is as described in the next section.

\subsection{Neural network via feature fusion}

The practices of physicians rely heavily on the synthesis of data from multiple sources. This includes, but is not limited to, structured laboratory data, unstructured text data, and imaging pixel data. Therefore, automated predictive models that can successfully utilize multimodal data may lead to better performance.

\begin{figure}[!htbp]
    \centering
    \includegraphics[width=\textwidth,trim=0 9em 0 0,clip]{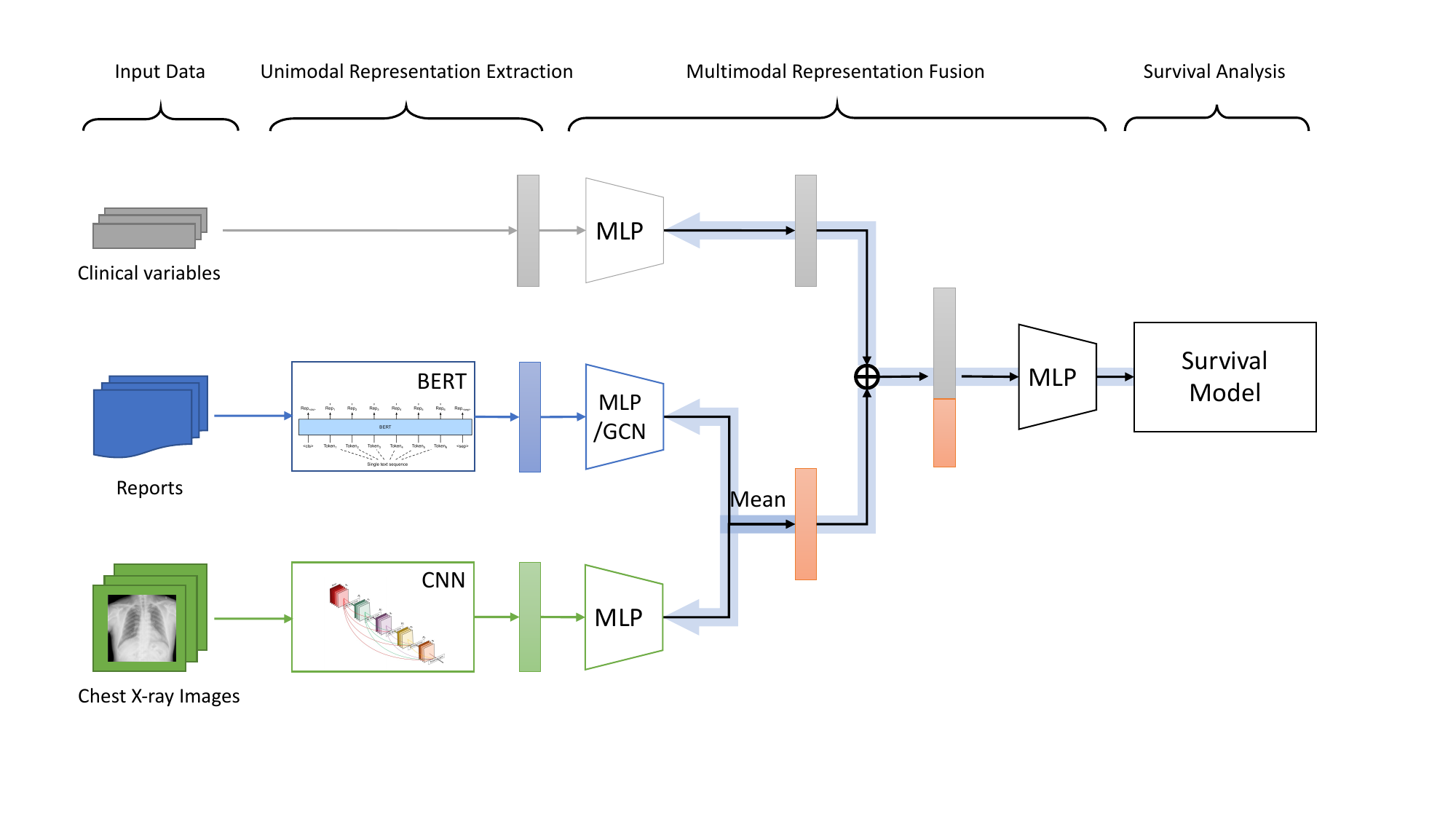}
    \caption{Multimodal feature fusion network.}
    \label{fig:flowchart}
\end{figure}

In this paper, we expand $\psi(x_{i})$ by introducing a deep neural network with the fusion features from multiple sources: SAPS-II risk factors $x_{saps}$, text features $x_{text}$, and imaging features $x_{img}$ (Figure~\ref{fig:flowchart}). The extracted text features $x_{text}$ and image features $x_{img}$ are respectively passed to two separate Multilayer Perceptron (MLP) modules where the output dimensions are equal. We then use the two hidden features by element-wise averaging. Finally, we concatenate it to $x_{saps}$.
\begin{equation}
    x_i = Avg(\text{DNN}_{img}(x_{img}), \text{DNN}_{text}(x_{text})) \oplus x_{saps}
\end{equation}

In terms of fusion strategy, our approach is similar to ``early fusion'' which refers to the process of combining features from multiple input modalities into one single feature vector before feeding into the survival model~\cite{huang2020fusion}. The difference is that our loss is propagated back to the DNNs during training, thus creating better feature selections for each training iteration. In addition, our approach is not ``joint fusion'' because the parameters of the features are not updated during the training iteration.

\subsection{Feature Extraction}

\subsubsection{SAPS-II score and risk factors}

SAPS-II was designed to measure the severity of disease for patients aged 18 or more admitted to ICU~\cite{le1993new}. Twenty-four hours after admission to the ICU, the measurements have been completed and resulted in an integer point score between 0 and 163. The score is calculated from 15 routine physiological measurements including information about previous health status, and some information obtained at admission. These measurements are: Age, Heart rate, Blood pressure, Temperature, PaO$_{2}$/FiO$_{2}$, Blood urea nitrogen, Urine output, Sodium, Potassium, Bicarbonate, Bilirubin, White blood count, Glasgow Coma Scale, Chronic disease, and Admission type.

\hypertarget{text-features}{%
\subsubsection{Text Features}\label{text-features}}

In this work, we investigate three sets of text features.

\textbf{Common thorax diseases from radiology reports}. The first set of features consists of 13 pre-defined diseases commonly found in radiology reports (Atelectasis, Cardiomegaly, Consolidation, Edema, Enlarged Cardiomediastinum, Fracture, Lung lesion, Lung opacity, Pleural effusion, Pleural other, Pneumonia, Pneumothorax, Support devices) and Normal~\cite{wang2017chestxray8, irvin2019chexpert, johnson2019mimiccxrjpg}. These labels were extracted from radiology reports using NegBio~\cite{peng2018negbio} and can be obtained from the MIMIC-CXR website\footnote{\url{https://physionet.org/content/mimic-cxr-jpg/2.0.0/}}.

\textbf{Transformer-based features}. The second set of features are text embeddings extracted by the BERT model - taking advantage of the pre-training on large-scale biomedical and clinical text corpora. Utilizing clinical texts in survival analysis is difficult because they are largely unstructured. The above lung diseases may fail to capture the textual information comprehensively since their labels are still limited in scope. In this work, we use the BERT-based hidden layer representations as text features. For an input report that contains $m$ tokens, the BERT model will produce a $d$-dimension embedding vector for each token, resulting in an $m \times d$ representation vector of the report in the latent space. We apply average pooling over the token embeddings from the last layer of the BERT model to obtain an aggregate latent report representation.

\textbf{GCN-based features}. We build a graph convolutional neural network (GCN) to model the inner correlations among radiology concepts. The graph was manually defined by domain experts (Figure 1 in Zhang et al.~\cite{Zhang2020WhenRR}. Disease findings are defined as nodes in the graph, and correlated findings are connected to influence each other during graph propagation. We take the $m \times d$ hidden representation vectors from the last layer of the BERT model. To initialize GCN node features, we apply a 1-dimension convolution over the text features with the kernel size $k$ and the number of output channels equal to the number of graph nodes. In this way, the graph nodes are initialized by aggregating the hidden features of all the tokens in the report.

The GCN updates its node representations by message passing. We first calculate $\hat{A} = D^{-1/2}\tilde{A}D^{-1/2}$ in a pre-processing step. $\tilde{A}$ = $A$ + $I_{N}$ is the adjacency matrix with added self-connections, where $A$ is the graph adjacency matrix, $I_{N}$ is the $N$-dimension identity matrix, $D = \text{diag}\sum_{j}^{}A_{ij}$ is the diagonal node degree matrix. Then based on~\cite{Kipf:2016tc}, the graph convolution can be expressed as:
\begin{align*}
    H^l & = ReLu(\hat{A}H^{0}W^{0} + b^0)\\
    Z & = softmax(\hat{A}H^{1}W^{1} + b^1)
\end{align*}
where $H^{l}$ is the states in the $l$-th layer, with $H^{0}$ initialized using the aggregate report text hidden features, and $W^{l}$ is a trainable layer-specific weights matrix.

\subsubsection{Image Features}

For image feature extraction, we use the ChexNet, a DenseNet-121 model pretrained on CheXpert dataset~\cite{irvin2019chexpert, huang2017densely, rajpurkar2017chexnet}. For each input image, we extract the image features of dimension $d_{img}$ from the global average pooling layer of DenseNet-121.

\section{Experiments}

\subsection{Study population and patient selection}

We use the MIMIC-IV dataset (Medical Information Mart for Intensive Care IV) to evaluate the proposed model~\cite{johnsonalistairmimiciv}. MIMIC-IV is a de-identified clinical database composed of 382,278 patients admitted in the ICUs at Beth Israel Deaconess Medical Center. Of those, we excluded patients who had no CXR studies before the measurements have been completed and resulted in the SAPS-II score. Therefore, there are in total 9,928 patients included in this study. Out of these patients, 2,213 patients (22\%) were deceased in the ICU. Table~\ref{demo_ICU} lists the information of the ICU admission group studied in this work. Details of the SAPS-II can be found in Table~\ref{saps_ICU}.

\begin{table}
\centering
\caption{\label{demo_ICU}
Information of the ICU admission group. EC: Enlarged Cardiomediastinum.}
\begin{tabular}{@{}lrrr@{}}
\toprule
& \textbf{ICU Discharge} & \textbf{ICU morality}\\
\midrule
Patient, n & 7,715 & 2,213\\
Age, mean (SD), y & 61.78 (18.20) & 69.63 (14.84) \\
Gender, male/female, \% & 45/55 & 46/54 \\
Race, \% & & \\
\hspace{1em}American Indian & 0.27 & 0.18 \\
\hspace{1em}Asian & 3.25 & 0.39 \\
\hspace{1em}Black/African American & 11.43 & 11.16 \\
\hspace{1em}Hispanic/Latino & 3.88 & 2.67 \\
\hspace{1em}White & 65.37 & 63.76 \\
\hspace{1em}Other & 5.22 & 4.34 \\
\hspace{1em}Unable to obtain & 0.73 & 1.04 \\
\hspace{1em}Unknown & 9.85 & 12.97 \\
Common thorax diseases, \%\\
\hspace{1em}Atelectasis & 19.90 & 21.69 \\
\hspace{1em}Cardiomegaly & 17.64 & 18.08 \\
\hspace{1em}Consolidation & 4.19 & 8.00 \\
\hspace{1em}Edema & 13.45 & 19.52 \\
\hspace{1em}EC & 3.54 & 4.07 \\
\hspace{1em}Fracture & 2.68 & 2.21 \\
\hspace{1em}Lung Lesion & 2.64 & 4.97 \\
\hspace{1em}Lung Opacity & 26.36 & 36.69 \\
\hspace{1em}Pleural Effusion & 17.04 & 28.29 \\
\hspace{1em}Pleural Other & 0.62 & 0.72 \\
\hspace{1em}Pneumonia & 0.70 & 9.31 \\
\hspace{1em}Pneumothorax & 2.44 & 2.94 \\
\hspace{1em}Support Devices & 34.50 & 43.83 \\
\bottomrule
\end{tabular}
\end{table}

\subsection{Evaluation metrics}

To assess the accuracy of our models, we use the C-index, defined as:

\begin{align*}
{L}_{s} = \frac{\sum_{i,j}I(T_{i} \geq T_{j}) \cdot I(R_{i} \leq R_{j}) }{\sum_{i,j}I(T_{i} \geq T_{j}) \cdot d_{j}},
\end{align*}

\noindent
where 
$I(c) = \begin{cases}
1& \text{if c is true}\\
0& \text{otherwise}
\end{cases}$, $d_{j} = \begin{cases}
1& \text{if $T_{j}$ exist}\\
0& \text{otherwise}
\end{cases}$
, $j \in \{1, 2, \cdots, N\}$, and $j \textgreater i$. $N$ is the number of sample.

Intuitively, the C-index measures the extent to which the model can assign logical risk scores. An individual with a shorter time-to-event $T$ should have a higher risk score $R$ than those with a longer time-to-event. C-index assigns a random model 0.5 and a perfect model 1.

\subsection{Implementation and Experimental Settings}

We perform a grid search to find the optimal hyperparameters based on the metrics and use them for all configurations. The MLP layer for SAPS-II risk factors takes an input of 15 dimensions, and fully connects to 15 output dimensions. The MLP layer for labels features fully connects the 14-dimension inputs to the 14-dimension outputs. The MLP layer for report text features fully connects the 768-dimension inputs to the 32-dimension outputs, and the MLP layer for chest X-ray image features fully connects the 1024-dimension inputs to the 32-dimension outputs.

We use 200 bootstrap samples to obtain a distribution of the C-index and report the 95\% confidence intervals (CI). For each bootstrap experiment, we sample $n$ patients with replacement from the whole set of $n$ patients. We then split the sampled set into training (70\%), validation (10\%), and test (20\%) sets. We iterate the training process for 250 epochs with batch size 72 and early stop if the validation loss does not decrease. The dropout rate is 0.5. The learning rate is 0.001 with an Adam optimizer~\cite{kingma2014adam}.

We obtained the SAPS-II scores using the scripts in the MIMIC-IV repository\footnote{\url{https://github.com/MIT-LCP/mimic-iv}}. The text embeddings are extracted using BlueBERT~\cite{peng2019transfer}, which was pre-trained on the PubMed abstracts and MIMIC-III notes. We use pycox\footnote{\url{https://github.com/havakv/pycox}}, scikit-survival~\cite{polsterl2020scikit}, and PyTorch to implement the framework. Intel Core i9-9960X 16 cores processor and NVIDIA Quadro RTX 5000 GPU are used in this work.

\section{Results}\label{results}

We first compare the baseline ICU scoring model and our models with four different feature settings. The SAPS-II score is an integer point score between 0 and 163 directly obtained from the MIMIC-IV website. The SAPS-II risk factors model is trained using the 15 routine physiological measurements. The SAPS-II risk factors + GCN features model is enriched with the GCN-based features. The SAPS-II risk factors + Image features model is enriched with chest X-ray image features. The multimodal features model is trained using SAPS-II risk factors combined with text features and chest X-ray image features using early average fusion.

Table \ref{single-column-result-1} shows that the ICU scoring model achieves an average C-index of 0.7470 (95\% confidence interval, 0.7263-0.7676). The mean C-index of our model with SAPS-II risk factors achieves 0.7545 (0.7240-0.7849), which brings 0.75\% improvements to the ICU scoring baseline model. When combining the SAPS-II risk factors with GCN-based text features and image features, the models obtain the average C-index of 0.7720 (0.7517-0.7923) and 0.7752 (0.7518-0.7985), respectively, yielding increases of 2.50\% and 2.82\%. Using the multimodal features, the performance of the model can further be boosted. We obtain the average C-index of 0.7829 (0.7620-0.8038), resulting in an improvement of 3.60\% over the ICU scoring model. We also train the multimodal features model with SAPS-II risk factors combined with GCN features and chest X-ray image features using early average fusion. The average C-index is 0.7805 (0.7570-0.8040), which is slightly lower than the proposed multimodal features model.

\begin{table}[!htbp]
\centering
\caption{\label{single-column-result-1}
C-index comparisons of the models using different sets of features.}
\begin{tabular}{lrr}
\toprule
Model     & C-index & (95\% CI)\\
\midrule
SAPS-II scores (ICU scoring baseline)        & 0.7477 & (0.7238-0.7716)  \\
SAPS-II risk factors     & 0.7555 & (0.7220-0.7890) \\
SAPS-II risk factors + GCN features  & 0.7745 & (0.7486-0.8004) \\
SAPS-II risk factors + Image features      & 0.7757 & (0.7522-0.7992) \\
Multimodal features & \textbf{0.7847} & (0.7625-0.8068)  \\
\bottomrule
\end{tabular}
\end{table}

\begin{figure}[!htbp]
    \centering
    \includegraphics[width=.9\textwidth]{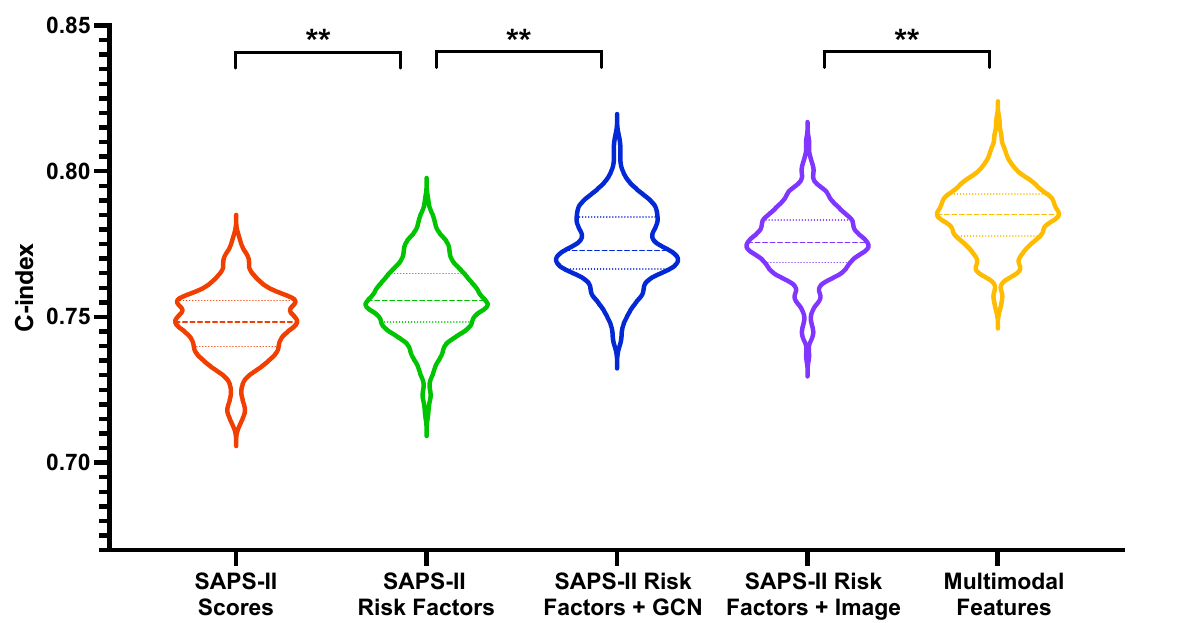}
    \caption{C-index comparisons of the models using different sets of features. 
    $*$: $p-$value $\leq$ 0.05; $**$: $p-$value $\leq$ 0.01.}
    \label{fig:model-comparison}
\end{figure}

Figure~\ref{fig:model-comparison} shows more details on bootstrapping. The violin shape reflects the distribution of the C-index: the thicker, the higher frequency. We find that the average C-index associated with the multimodal features model is statistically higher than the other four settings.

\begin{figure}[!htbp]
    \centering
    \includegraphics[width=.7\columnwidth]{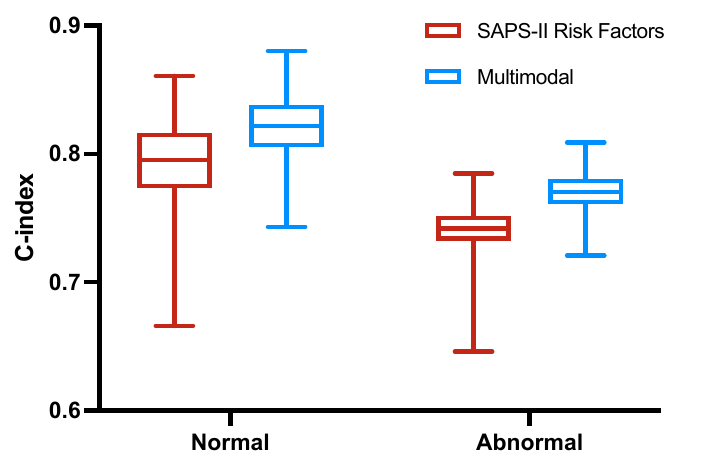}
    \caption{The C-index results of the models trained on the entire dataset and tested on the normal patients or patients with chest X-ray abnormalities.}
    \label{fig:normal}
\end{figure}

\begin{figure}[!htbp]
    \centering
    \includegraphics[width=\textwidth]{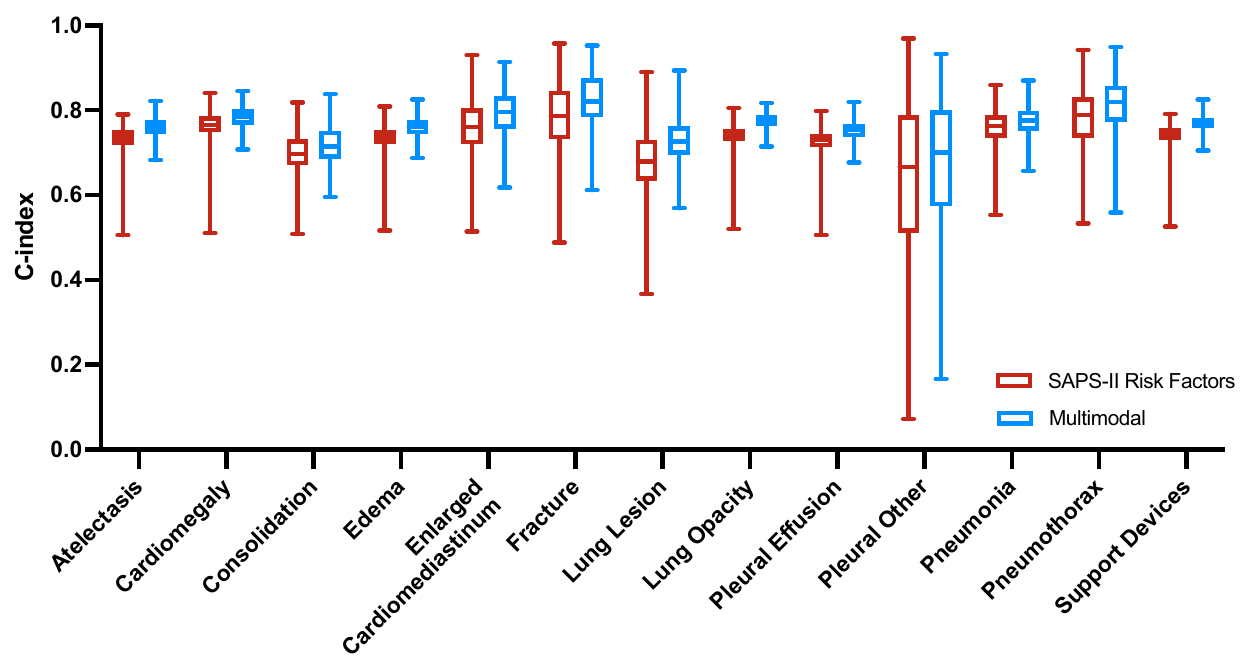}
    \caption{The C-index results of the models trained on the entire dataset and tested on the patients with different chest X-ray abnormalities.}
    \label{fig:abnormalities}
\end{figure}

Figure \ref{fig:normal} shows the C-index results of our SAPS-II risk factors model and multimodal features model, marked in red and blue respectively. Both are trained on the entire dataset and tested on the patients with normal or abnormal chest X-rays. It is clear that our multimodal features model outperforms the SAPS-II risk factors model and the normal subjects can be more accurately predicted by our model.
Figure \ref{fig:abnormalities} further breaks chest X-ray abnormalities into 13 pre-defined thorax diseases.


\section{Discussion}

\subsection{Comparison of different types of text features.}

First, we compare the results of our model using different types of text features. SAPS-II risk factors + labels, SAPS-II risk factors + transformer features, and SAPS-II risk factors + GCN features. They are trained using 15 routine physiological measurements combined with 14 thorax disease labels, transformer-based features, and GCN-based features, respectively. Table \ref{tab:text features} lists the results of our model using these three feature settings. The mean C-indexes for these three settings are 0.7669 (0.7456-0.7882), 0.7714 (0.7488-0.7941), and 0.7720 (0.7517-0.7923), respectively. Models with transformer features or GCN features outperform the model that uses only labels. But there is no significant difference between the transformer and GCN features.

\begin{table}[!htpb]
\centering
\caption{\label{tab:text features}
The C-index results of the models using different types of text features.}
\begin{tabular}{lrr}
\toprule
Model     & C-index & (95\% CI)\\
\midrule
SAPS-II risk factors + labels       & 0.7669 & (0.7456-0.7882) \\
SAPS-II risk factors + transformer features & 0.7714 & (0.7488-0.7941) \\
SAPS-II risk factors + GCN features  & \textbf{0.7720} & (0.7517-0.7923) \\
\bottomrule
\end{tabular}
\end{table}

\subsection{Contribution of thorax diseases in survival analysis}

Next, we analyze the multivariate association of chest X-ray abnormalities to ICU mortality based on Cox Proportion Hazards (CoxPH model) (Table \ref{table:feature_weights}). The p-values of these four findings: enlarged cardiomediastinum, fracture, pneumonia, pneumothorax are greater than 0.05, indicating that there is no statistically significant difference. In other words, these findings do not contribute to mortality prediction.

\begin{table}[!htbp]
\caption{\label{table:feature_weights}
Multivariate associations of chest X-ray abnormalities to ICU-mortality.
*: $p-$value $\leq$ 0.05; **: $p-$value $\leq$ 0.01; ***: $p-$value $<$ 0.001.}
\begin{tabular}{lcr@{--}lc}
\toprule
                            & Hazard ratio & \multicolumn{2}{c}{95\% CI} & $p$-value \\
\midrule
Atelectasis                 & 0.84         & 0.75         & 0.94         & **      \\
Cardiomegaly                & 0.85         & 0.76         & 0.96         & **      \\
Consolidation               & 1.33         & 1.14         & 1.55         & ***     \\
Edema                       & 1.23         & 1.10         & 1.38         & ***     \\
Enlarged  Cardiomediastinum & 0.91         & 0.75         & 1.12         & 0.37    \\
Fracture                    & 0.96         & 0.72         & 1.28         & 0.77    \\
Lung  Lesion                & 1.37         & 1.13         & 1.67         & **      \\
Lung  Opacity               & 1.29         & 1.17         & 1.42         & ***     \\
Pleural  Effusion           & 1.13         & 1.02         & 1.26         & *       \\
Pleural  Other              & 0.64         & 0.41         & 1.00         & *       \\
Pneumonia                   & 1.07         & 0.93         & 1.23         & 0.34    \\
Pneumothorax                & 1.10         & 0.86         & 1.41         & 0.45    \\
Support  Devices            & 1.27         & 1.16         & 1.39         & ***    \\
\bottomrule
\end{tabular}
\end{table}

\subsection{Comparison of linear and deep survival models}

We then compare the performances of the linear machine learning model and deep learning model: CoxPH {[}40{]} and DeepSurv-based model.

Table \ref{tab:model} shows the results for both models with two feature settings. The average C-indexes of the CoxPH model with SAPS-II risk factors and SAPS-II risk factors + labels are 0.7510 (0.7300-0.7720) and 0.7617 (0.7414-0.7819), respectively, in comparison with 0.7545 (0.7240-0.7849) and 0.7669 (0.7456-0.7882) obtained by our DeepSurv-based model. The results demonstrate that deep learning models outperform CoxPH on high-dimensional features. The p-value for CoxPH and DeepSurv-based model using SAPS-II is 0.01 and p-value is 1.08e-6 when using SAPS-II + labels.

\begin{table}[!htbp]
\centering
\caption{\label{tab:model}
The C-index results of the conventional machine learning models and the deep learning models trained and tested on the entire dataset.}
\begin{tabular}{lrr}
\toprule
Model     & C-index & (95\% CI)\\
\midrule
SAPS-II risk factors \\
\hspace{2em}CoxPH     & 0.7510 & (0.7300-0.7720) \\
\hspace{2em}DeepSurv-based  & \textbf{0.7545} & (0.7240-0.7849) \\
SAPS-II risk factors + labels   \\
\hspace{2em}CoxPH     & 0.7617 & (0.7414-0.7819) \\
\hspace{2em}DeepSurv-based  & \textbf{0.7669} & (0.7456-0.7882) \\
\bottomrule
\end{tabular}
\end{table}

\subsection{Error Analysis}

\begin{figure}[!htbp]
    \centering
    \includegraphics[width=.8\columnwidth]{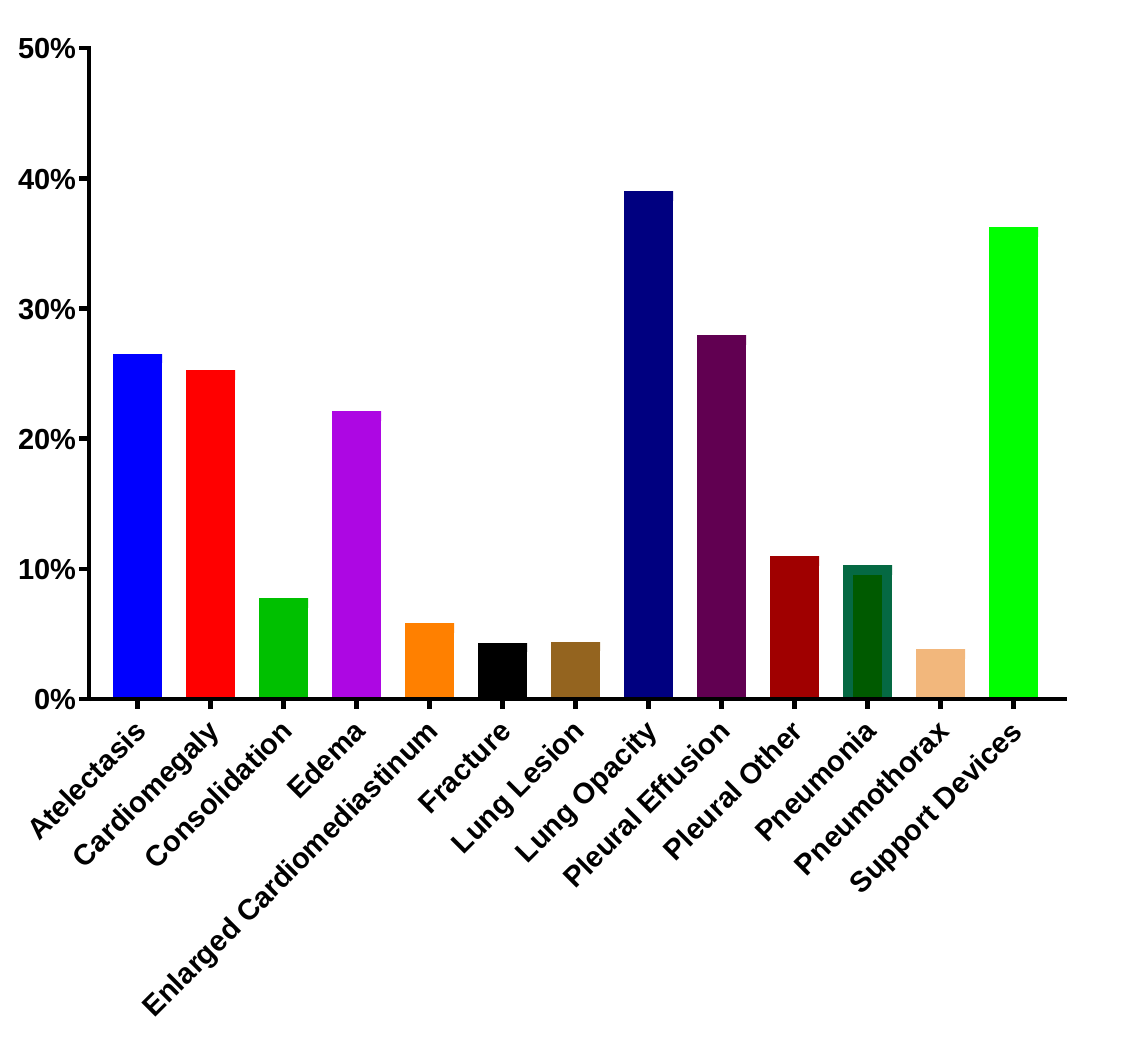}
    \caption{Distribution of thorax diseases among patients where our multimodal model made more accurate predictions than SAPS-II.}
    \label{fig:case_study}
\end{figure}

Error analysis (i.e., examining the reasons behind inaccurate predictions) revealed that the multimodal accounted for fewer errors. Table~\ref{{table:case_study_examples}} demonstrates one example case of ICU mortality. According to physiological measurements, SAPS-II graded patient \#1 the score of 38 and patient \#2 36. However, patient \#1 was decreased at hour 198, but patient \#2 was deceased at hour 75. Hence, the SAPS-II incorrectly assigned the score. However, our multimodal approach correctly assigned a higher survival probability to patient \#1 (0.9903) than to patient \#2 (0.9562). In one bootstrap sample, we observed a total of 40,529 such errors (patient \#1 has normal chest x-ray, and SAPS-II gives wrong predictions but our multimodal method gives correct predicitons) with 1,802 distinct patients, out of which 527 patients have normal chest X-rays and 1,275 patients have abnormal chest x-rays. Figure~\ref{fig:case_study} shows the distribution of thorax diseases among 1,275 patients. It shows that Lung Opacity (38.98\%) contributes most to the ICU mortality prediction.

\section{Conclusion}

In this paper, we propose a deep learning method that combines text and image features that may not be found in the structured EHR field before, to improve the ICU mortality prediction. Experiments on the MIMIC-IV dataset show that our multimodal features model (using SAPS-II risk factors and early fusion of text and X-ray image features) obtains a superior average C-index of 0.7829 (0.7620-0.8038) than several baselines. This demonstrates that the additional information provided by the multimodal features can improve the ICU mortality prediction performance. We also investigate whether deep learning methods are more powerful than traditional machine learning methods in predicting ICU mortality. Through experiments, our model achieves a better average C-index than the CoxPH model under the same feature fusion setting, proving the superior performance of deep learning methods in ICU mortality prediction.

There are several limitations to this work. First, we use a fusion strategy similar to ``early fusion'' to fuse the text and image features extracted by BlueBERT and ChexNet, respectively, but the parameters of them are not updated during the training iterations. In the future, we plan to use joint fusion to propagate the loss back to the feature extraction modules during training, which may improve the representation learning performance. Second, a knowledge graph is a popular tool for representing background knowledge, which can improve several aspects of the model. We will explore other domain knowledge and try different ways of incorporating the knowledge graph into ICU mortality prediction. Third, the longitudinal EHR data contain the information regarding the disease progressions that may help ICU mortality prediction, but are not utilized in this work. In the future, we can employ the longitudinal EHR to assist in predicting ICU mortality. Fourth, there is a risk of selction bias in this study. For instance, we only included in our analysis patients with imaging studies after ICU admission. Imaging studies are usually performed when a patient is sicker, for example, to confirm central line placement. This selection could lead to a sample that is not representative of the ICU population. However, selection bias is a common problem in machine learning~\cite{cortes2008sample}, statistics~\cite{whittemore1978collapsibility}, and epidemiology~\cite{robins2001data}; as a result, a number of techniques have been developed to correct it. In the future, we will investigate these techniques. Fifth, machine learning models are vulnerable to adversarial attacks~\cite{goodfellow2014explaining}. For example, images can be attacked by adding a small perturbation to the original images. Texts can be attacked by adding a small number of words. These attacks are imperceptible to humans but mislead a model into producing incorrect outputs. Like selection bias, the adversarial attack is a common problem in the medical domain, where accurate diagnostic results are of paramount importance~\cite{paschali2018generalizability}. Previous studies suggest that if a model could eliminate noises in their learned feature representations, they would be more robust against adversarial perturbations~\cite{finlayson2019adversarial}. We will study these techniques to improve the robustness of the model in the future.

While our work only scratches the surface of multimodal fusion for survival analysis, we hope it will shed light on the future directions for ICU mortality prediction.

\backmatter

\bmhead{Supplementary information}

\bmhead{Abbreviations}

\begin{itemize}
   \item ICU: Intensive Care Unit
   \item EHR: Electronic Health Record
   \item SAPS: Simplified Acute Physiology Score
   \item MIMIC: Medical Information Mart for Intensive Care
   \item NLP: Natural Language Processing
   \item GP: Gaussian Process
   \item CNN: Convolutional Neural Network
   \item MLP: Multilayer Perceptron
   \item GCN: Graph Convolutional Network
   \item CI: Confidence intervals
\end{itemize}




\section*{Declarations}

\bmhead{Ethics approval and consent to participate}

The dataset supporting the conclusions of this article is available in the Medical Information Mart for Intensive Care version IV (MIMIC-IV), which is a public de-identified database thus informed consent and approval of the Institutional Review Board was waived. Our access to the database was approved after completion of the Collaborative Institutional Training Initiative (CITI program) web-based training.

\bmhead{Consent for publication}

N/A

\bmhead{Availability of data and materials}

We made our codes publicly available at \url{https://github.com/bionlplab/mimic-icu-mortality}. The dataset we are using in this work is Medical Information Mart for Intensive Care IV (MIMIC-IV), which is also publicly available at \url{https://physionet.org/content/mimiciv/0.4/}.

\bmhead{Competing interests}

The authors declare that they have no competing interests.

\bmhead{Funding}

This work was funded by the National Library of Medicine under award number 4R00LM013001 and Amazon Machine Learning Grant.

\bmhead{Author's contributions}

ML and SW implemented the methods, conducted the experiments and wrote the paper. YP advised on all aspects of the work involved in this project and assisted in the paper writing. YD, LZ, and FW advised on the overall direction of the project and edited the paper. All authors read and approved the final manuscript.

\begin{appendices}

\section{Support materials}\label{secA1}





\newcommand{\fourcolumns}[1]{\multicolumn{4}{l}{#1}}
\begin{longtable}{lrrrr}
\caption{\label{saps_ICU}SAPS-II physiological measurements of the ICU admission group.}
\\
\toprule
 & \textbf{Score} & \textbf{ICU Discharge} & \textbf{ICU Morality}\\
 && \% & \%\\
\midrule
\endfirsthead
\multicolumn{4}{r}{\small\sl continued from previous page}\\
\toprule
 & \textbf{Score} & \textbf{ICU Discharge} & \textbf{ICU Morality}\\
 && \% & \% \\
\midrule
\endhead
\bottomrule
\multicolumn{4}{r}{\small\sl continued on next page}\\
\endfoot
\bottomrule
\endlastfoot
\fourcolumns{Age, year}\\
\hspace{1em} $<$40   & 0  & 12.51 & 3.21\\
\hspace{1em} 40-59   & 7  & 25.91 & 17.67\\
\hspace{1em} 60-69   & 12 & 21.63 & 20.70\\
\hspace{1em} 70-74   & 15 & 9.16  & 11.84\\
\hspace{1em} 75-79   & 16 & 9.19  & 11.21\\
\hspace{1em} $\ge$80 & 18 & 21.59 & 35.38\\

\fourcolumns{Heart rate}\\
\hspace{1em} $<$40    & 11 & 21.04 & 31.22\\
\hspace{1em} 40-69    & 2  & 42.28 & 28.11\\
\hspace{1em} 70-119   & 0  & 34.71 & 32.90\\
\hspace{1em} 120-159  & 4  & 0.93  & 2.94\\
\hspace{1em} $\ge$160 & 7  & 1.04  & 4.84\\

\fourcolumns{Systolic BP, mmHg}\\
\hspace{1em} $<$70    & 13 & 5.51  & 22.01\\
\hspace{1em} 70-99    & 5  & 62.49 & 59.51\\
\hspace{1em} 100-199  & 0  & 30.82 & 17.44\\
\hspace{1em} $\ge$200 & 2  & 1.18  & 1.04\\

\fourcolumns{Temperature $\le 39$\textdegree C}\\
\hspace{1em} No  & 0 & 95.58 & 93.76\\
\hspace{1em} Yes & 3 & 4.42  & 6.24 \\

\fourcolumns{PaO$_2$/FiO$_2$, mm Hg}\\
\hspace{1em} $<$100         & 11  & 3.47 & 13.29\\
\hspace{1em} 100-199        & 9  & 7.57 & 13.92\\
\hspace{1em} $\ge$200       & 6  & 13.66  & 17.40\\
\hspace{1em} No ventilation & 0  & 75.29  & 55.40\\

\fourcolumns{Blood urea nitrogen, mg/dL}\\
\hspace{1em} $<$28         & 0  & 70.28 & 47.40\\
\hspace{1em} 28-93         & 6  & 26.80 & 46.18\\
\hspace{1em} $\ge$84       & 10 & 2.92  & 6.42\\

\fourcolumns{Urine output, mL/day}\\
\hspace{1em} $<$500    & 11 & 7.66  & 29.60\\
\hspace{1em} 500-999   & 4  & 16.84 & 19.84\\
\hspace{1em} $\ge$1000 & 0  & 75.50 & 50.56\\

\fourcolumns{Sodium, mEq/L}\\
\hspace{1em} $<$125    & 5 & 2.20  & 2.58\\
\hspace{1em} 125-144   & 0 & 89.31 & 81.16\\
\hspace{1em} $\ge$145  & 1 & 8.49  & 16.27\\

\multicolumn{4}{l}{Potassium}\\
\hspace{1em} 3.0-4.9 & 0  & 82.31 & 69.95\\
\hspace{1em} $<$3.0 or $\ge$5.0 & 3 & 17.69 & 30.05 \\

\fourcolumns{Bicarbonate, mEq/L}\\
\hspace{1em} $<$15   & 6 & 4.72 & 16.49 \\
\hspace{1em} 15-19   & 3 & 18.09 & 26.25\\
\hspace{1em} $\ge$20 & 0 & 77.19 & 67.25\\

\fourcolumns{Bilirubin, mg/dL}\\
\hspace{1em} $<$4.0 & 0  & 96.64 & 90.24\\
\hspace{1em} 4.0-5.9 & 4 & 1.24 & 2.76\\
\hspace{1em} $\ge$6.0 & 9  & 2.11  & 7.00\\

\multicolumn{4}{l}{White blood count, x10$^3$/mm$^3$}\\
\hspace{1em} $<$1.0 & 12 & 0.45 & 2.08 \\
\hspace{1em} 1.0-19.9 & 0 & 89.38 & 77.09\\
\hspace{1em} $\ge$20.0 & 3 & 10.16 & 20.83\\

\fourcolumns{Glasgow Coma Scale}\\
\hspace{1em} 14-15 & 0  & 78.70 & 68.59\\
\hspace{1em} 11-13 & 5  & 11.67 & 10.94\\
\hspace{1em} 9-10  & 7  & 3.69  & 9.47\\
\hspace{1em} 6-8   & 13 & 3.85  & 7.14\\
\hspace{1em} $<$6  & 26 & 2.09  & 7.86\\

\fourcolumns{Chronic disease}\\
\hspace{1em} None    & 0 & 90.56 & 76.59\\
\hspace{1em} Metastatic cancer   & 9 & 6.13 & 15.68\\
\hspace{1em} \footnotesize{Hematologic malignancy} & 10 & 2.40 & 7.05\\
\hspace{1em} AIDS & 17 & 0.91 & 0.68\\

\fourcolumns{Type of admission}\\
\hspace{1em} Scheduled surgical   & 0 & 0.75  & 0.09\\
\hspace{1em} Medical              & 6 & 83.50 & 86.58\\
\hspace{1em} Unscheduled surgical & 8 & 15.75 & 13.33\\
\end{longtable}

\begin{table}
\centering
\caption{\label{table:case_study_examples}
An example case of ICU mortality prediction where our multimodal model made more accurate prediction than SAPS-II. According to physiological measurements, SAPS-II graded patient 1 the score of 38 and patient 2 36. However, patient 1 was decreased at hour 198, but patient 2 was deceased at hour 75. Hence, the SAPS-II \underline{incorrectly} assigned the score. However, our multimodal approach correctly assigned a higher survival probability to patient 1 (0.9903) to patient 2 (0.9562).}
\begin{tabular}{m{2.6cm}p{5cm}p{5cm}}
\toprule
 & \textbf{Patient 1} & \textbf{Patient 2}\\
\midrule
Age, y & 85 & 62\\
Gender & Male & Male\\
Race & African American & White\\
Time-to-event, hr &  198 & 75 \\
Mortality score & 38 & 36 \\
Findings & No Findings & Atelectasis, Consolidation, Lung Lesion, Pleural Effusion \\
 \midrule
Report & 
FINDINGS:  Portable AP chest radiograph demonstrates no focal consolidation,
 pleural effusion, pulmonary vascular engorgement, or pneumothorax.  Multiple
 vascular stents are noted in the right upper extremity, superior mediastinum,
 and mid left upper extremity. The aorta is tortuous. The cardiomediastinal
 silhouette is otherwise normal.
 
IMPRESSION:  No acute cardiopulmonary process. & 
FINDINGS:  There continues to be near-complete opacification of the right lung
 compatible with components of pleural effusion and post-obstructive
 consolidation secondary to a known right chest mass.  Additionally, there is
 increasing left pleural effusion with basilar atelectasis.  Assessment of the
 heart size is limited due to these opacities. There is no pneumothorax.
 
IMPRESSION:  Stable appearance of right-sided pleural effusion and
 post-obstructive consolidation in the setting of a known right chest mass;
 increasing left pleural effusion with basal atelectasis.\\
  \midrule
Chest X-ray &
\adjustbox{valign=t}{ \includegraphics[width=5cm]{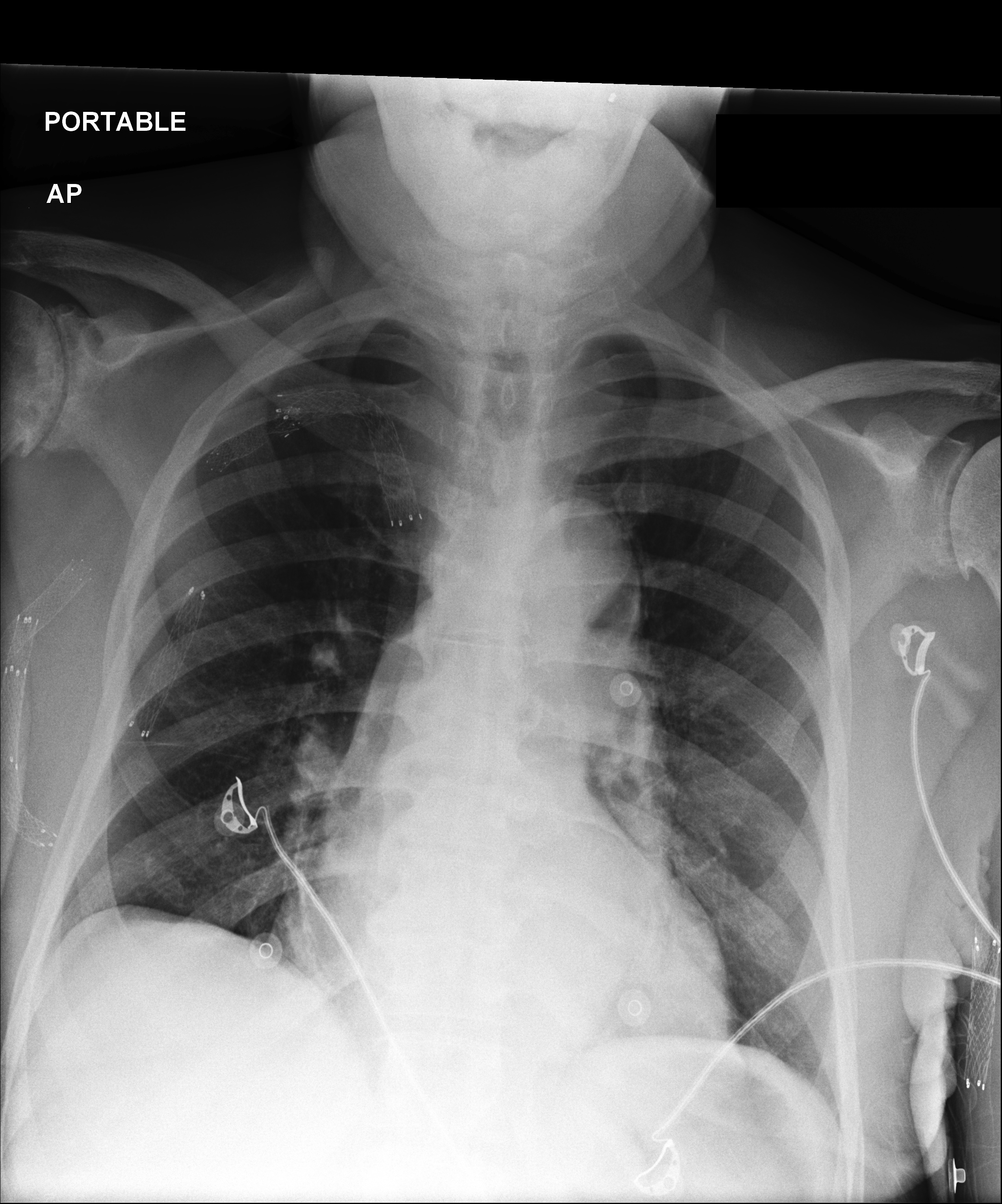}}
 & 
 \adjustbox{valign=t}{\includegraphics[width=5cm]{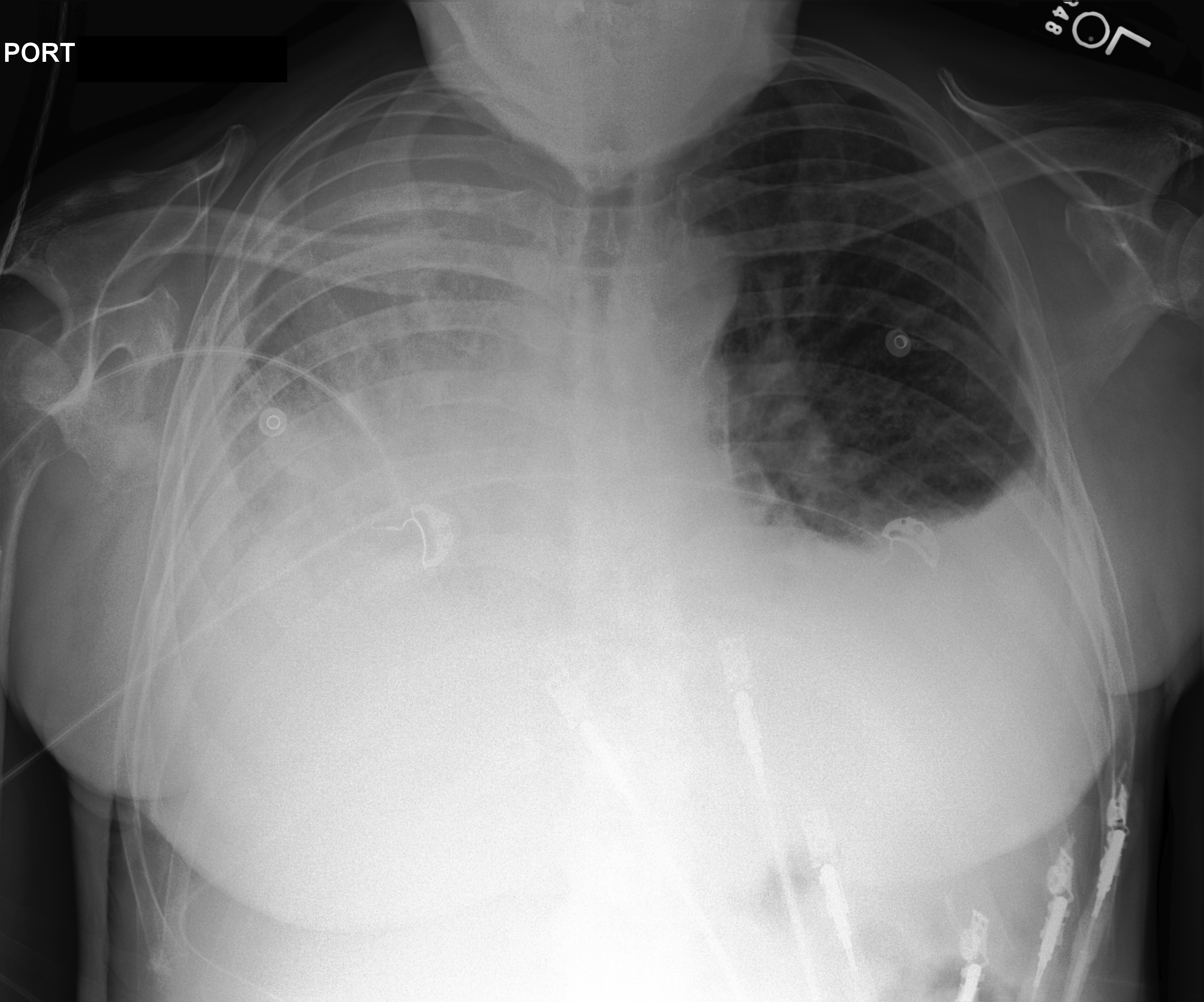}}\\
\bottomrule
\end{tabular}
\end{table}

\end{appendices}


\bibliography{sn-bibliography}


\begin{thebibliography}{44}
\ifx \bisbn   \undefined \def \bisbn  #1{ISBN #1}\fi
\ifx \binits  \undefined \def \binits#1{#1}\fi
\ifx \bauthor  \undefined \def \bauthor#1{#1}\fi
\ifx \batitle  \undefined \def \batitle#1{#1}\fi
\ifx \bjtitle  \undefined \def \bjtitle#1{#1}\fi
\ifx \bvolume  \undefined \def \bvolume#1{\textbf{#1}}\fi
\ifx \byear  \undefined \def \byear#1{#1}\fi
\ifx \bissue  \undefined \def \bissue#1{#1}\fi
\ifx \bfpage  \undefined \def \bfpage#1{#1}\fi
\ifx \blpage  \undefined \def \blpage #1{#1}\fi
\ifx \burl  \undefined \def \burl#1{\textsf{#1}}\fi
\ifx \doiurl  \undefined \def \doiurl#1{\url{https://doi.org/#1}}\fi
\ifx \betal  \undefined \def \betal{\textit{et al.}}\fi
\ifx \binstitute  \undefined \def \binstitute#1{#1}\fi
\ifx \binstitutionaled  \undefined \def \binstitutionaled#1{#1}\fi
\ifx \bctitle  \undefined \def \bctitle#1{#1}\fi
\ifx \beditor  \undefined \def \beditor#1{#1}\fi
\ifx \bpublisher  \undefined \def \bpublisher#1{#1}\fi
\ifx \bbtitle  \undefined \def \bbtitle#1{#1}\fi
\ifx \bedition  \undefined \def \bedition#1{#1}\fi
\ifx \bseriesno  \undefined \def \bseriesno#1{#1}\fi
\ifx \blocation  \undefined \def \blocation#1{#1}\fi
\ifx \bsertitle  \undefined \def \bsertitle#1{#1}\fi
\ifx \bsnm \undefined \def \bsnm#1{#1}\fi
\ifx \bsuffix \undefined \def \bsuffix#1{#1}\fi
\ifx \bparticle \undefined \def \bparticle#1{#1}\fi
\ifx \barticle \undefined \def \barticle#1{#1}\fi
\bibcommenthead
\ifx \bconfdate \undefined \def \bconfdate #1{#1}\fi
\ifx \botherref \undefined \def \botherref #1{#1}\fi
\ifx \url \undefined \def \url#1{\textsf{#1}}\fi
\ifx \bchapter \undefined \def \bchapter#1{#1}\fi
\ifx \bbook \undefined \def \bbook#1{#1}\fi
\ifx \bcomment \undefined \def \bcomment#1{#1}\fi
\ifx \oauthor \undefined \def \oauthor#1{#1}\fi
\ifx \citeauthoryear \undefined \def \citeauthoryear#1{#1}\fi
\ifx \endbibitem  \undefined \def \endbibitem {}\fi
\ifx \bconflocation  \undefined \def \bconflocation#1{#1}\fi
\ifx \arxivurl  \undefined \def \arxivurl#1{\textsf{#1}}\fi
\csname PreBibitemsHook\endcsname

\bibitem[\protect\citeauthoryear{Lipshutz
  et~al.}{2016}]{lipshutz2016predicting}
\begin{barticle}
\bauthor{\bsnm{Lipshutz}, \binits{A.K.}},
\bauthor{\bsnm{Feiner}, \binits{J.R.}},
\bauthor{\bsnm{Grimes}, \binits{B.}},
\bauthor{\bsnm{Gropper}, \binits{M.A.}}:
\batitle{Predicting mortality in the intensive care unit: a comparison of the
  university health consortium expected probability of mortality and the
  mortality prediction model iii}.
\bjtitle{Journal of intensive care}
\bvolume{4}(\bissue{1}),
\bfpage{1}--\blpage{8}
(\byear{2016})
\end{barticle}
\endbibitem

\bibitem[\protect\citeauthoryear{Zimmerman et~al.}{2006}]{zimmerman2006acute}
\begin{barticle}
\bauthor{\bsnm{Zimmerman}, \binits{J.E.}},
\bauthor{\bsnm{Kramer}, \binits{A.A.}},
\bauthor{\bsnm{McNair}, \binits{D.S.}},
\bauthor{\bsnm{Malila}, \binits{F.M.}}:
\batitle{Acute physiology and chronic health evaluation (apache) iv: hospital
  mortality assessment for today’s critically ill patients}.
\bjtitle{Critical care medicine}
\bvolume{34}(\bissue{5}),
\bfpage{1297}--\blpage{1310}
(\byear{2006})
\end{barticle}
\endbibitem

\bibitem[\protect\citeauthoryear{Le~Gall et~al.}{1993}]{le1993new}
\begin{barticle}
\bauthor{\bsnm{Le~Gall}, \binits{J.-R.}},
\bauthor{\bsnm{Lemeshow}, \binits{S.}},
\bauthor{\bsnm{Saulnier}, \binits{F.}}:
\batitle{A new simplified acute physiology score (saps ii) based on a
  european/north american multicenter study}.
\bjtitle{JAMA}
\bvolume{270}(\bissue{24}),
\bfpage{2957}--\blpage{2963}
(\byear{1993})
\end{barticle}
\endbibitem

\bibitem[\protect\citeauthoryear{Teres et~al.}{1987}]{teres1987validation}
\begin{barticle}
\bauthor{\bsnm{Teres}, \binits{D.}},
\bauthor{\bsnm{Lemeshow}, \binits{S.}},
\bauthor{\bsnm{Avrunin}, \binits{J.S.}},
\bauthor{\bsnm{Pastides}, \binits{H.}}:
\batitle{Validation of the mortality prediction model for icu patients.}
\bjtitle{Critical care medicine}
\bvolume{15}(\bissue{3}),
\bfpage{208}--\blpage{213}
(\byear{1987})
\end{barticle}
\endbibitem

\bibitem[\protect\citeauthoryear{El-Rashidy et~al.}{2020}]{el2020intensive}
\begin{barticle}
\bauthor{\bsnm{El-Rashidy}, \binits{N.}},
\bauthor{\bsnm{El-Sappagh}, \binits{S.}},
\bauthor{\bsnm{Abuhmed}, \binits{T.}},
\bauthor{\bsnm{Abdelrazek}, \binits{S.}},
\bauthor{\bsnm{El-Bakry}, \binits{H.M.}}:
\batitle{Intensive care unit mortality prediction: An improved patient-specific
  stacking ensemble model}.
\bjtitle{IEEE Access}
\bvolume{8},
\bfpage{133541}--\blpage{133564}
(\byear{2020})
\end{barticle}
\endbibitem

\bibitem[\protect\citeauthoryear{Ghassemi
  et~al.}{2015}]{ghassemi2015multivariate}
\begin{bchapter}
\bauthor{\bsnm{Ghassemi}, \binits{M.}},
\bauthor{\bsnm{Pimentel}, \binits{M.}},
\bauthor{\bsnm{Naumann}, \binits{T.}},
\bauthor{\bsnm{Brennan}, \binits{T.}},
\bauthor{\bsnm{Clifton}, \binits{D.}},
\bauthor{\bsnm{Szolovits}, \binits{P.}},
\bauthor{\bsnm{Feng}, \binits{M.}}:
\bctitle{A multivariate timeseries modeling approach to severity of illness
  assessment and forecasting in icu with sparse, heterogeneous clinical data}.
In: \bbtitle{Proceedings of the AAAI Conference on Artificial Intelligence},
vol. \bseriesno{29}
(\byear{2015})
\end{bchapter}
\endbibitem

\bibitem[\protect\citeauthoryear{Zhao et~al.}{2020}]{zhao2020prediction}
\begin{barticle}
\bauthor{\bsnm{Zhao}, \binits{Z.}},
\bauthor{\bsnm{Chen}, \binits{A.}},
\bauthor{\bsnm{Hou}, \binits{W.}},
\bauthor{\bsnm{Graham}, \binits{J.M.}},
\bauthor{\bsnm{Li}, \binits{H.}},
\bauthor{\bsnm{Richman}, \binits{P.S.}},
\bauthor{\bsnm{Thode}, \binits{H.C.}},
\bauthor{\bsnm{Singer}, \binits{A.J.}},
\bauthor{\bsnm{Duong}, \binits{T.Q.}}:
\batitle{Prediction model and risk scores of icu admission and mortality in
  covid-19}.
\bjtitle{PloS one}
\bvolume{15}(\bissue{7}),
\bfpage{0236618}
(\byear{2020})
\end{barticle}
\endbibitem

\bibitem[\protect\citeauthoryear{Johnson and Mark}{2017}]{johnson2017real}
\begin{bchapter}
\bauthor{\bsnm{Johnson}, \binits{A.E.}},
\bauthor{\bsnm{Mark}, \binits{R.G.}}:
\bctitle{Real-time mortality prediction in the intensive care unit}.
In: \bbtitle{AMIA Annual Symposium Proceedings},
vol. \bseriesno{2017},
p. \bfpage{994}
(\byear{2017}).
\bcomment{American Medical Informatics Association}
\end{bchapter}
\endbibitem

\bibitem[\protect\citeauthoryear{Murdoch and
  Detsky}{2013}]{murdoch2013inevitable}
\begin{barticle}
\bauthor{\bsnm{Murdoch}, \binits{T.B.}},
\bauthor{\bsnm{Detsky}, \binits{A.S.}}:
\batitle{The inevitable application of big data to health care}.
\bjtitle{Jama}
\bvolume{309}(\bissue{13}),
\bfpage{1351}--\blpage{1352}
(\byear{2013})
\end{barticle}
\endbibitem

\bibitem[\protect\citeauthoryear{Ching et~al.}{2018}]{ching2018opportunities}
\begin{barticle}
\bauthor{\bsnm{Ching}, \binits{T.}},
\bauthor{\bsnm{Himmelstein}, \binits{D.S.}},
\bauthor{\bsnm{Beaulieu-Jones}, \binits{B.K.}},
\bauthor{\bsnm{Kalinin}, \binits{A.A.}},
\bauthor{\bsnm{Do}, \binits{B.T.}},
\bauthor{\bsnm{Way}, \binits{G.P.}},
\bauthor{\bsnm{Ferrero}, \binits{E.}},
\bauthor{\bsnm{Agapow}, \binits{P.-M.}},
\bauthor{\bsnm{Zietz}, \binits{M.}},
\bauthor{\bsnm{Hoffman}, \binits{M.M.}}, \betal:
\batitle{Opportunities and obstacles for deep learning in biology and
  medicine}.
\bjtitle{Journal of The Royal Society Interface}
\bvolume{15}(\bissue{141}),
\bfpage{20170387}
(\byear{2018})
\end{barticle}
\endbibitem

\bibitem[\protect\citeauthoryear{Mechcatie and
  Rosenberg}{2018}]{mechcatie2018nursing}
\begin{barticle}
\bauthor{\bsnm{Mechcatie}, \binits{E.}},
\bauthor{\bsnm{Rosenberg}, \binits{K.}}:
\batitle{Nursing notes are predictive of outcomes in icu patients}.
\bjtitle{AJN The American Journal of Nursing}
\bvolume{118}(\bissue{10}),
\bfpage{70}
(\byear{2018})
\end{barticle}
\endbibitem

\bibitem[\protect\citeauthoryear{Yang et~al.}{2021}]{yang2021multimodal}
\begin{barticle}
\bauthor{\bsnm{Yang}, \binits{H.}},
\bauthor{\bsnm{Kuang}, \binits{L.}},
\bauthor{\bsnm{Xia}, \binits{F.}}:
\batitle{Multimodal temporal-clinical note network for mortality prediction}.
\bjtitle{Journal of Biomedical Semantics}
\bvolume{12}(\bissue{1}),
\bfpage{1}--\blpage{14}
(\byear{2021})
\end{barticle}
\endbibitem

\bibitem[\protect\citeauthoryear{Grnarova et~al.}{2016}]{grnarova2016neural}
\begin{botherref}
\oauthor{\bsnm{Grnarova}, \binits{P.}},
\oauthor{\bsnm{Schmidt}, \binits{F.}},
\oauthor{\bsnm{Hyland}, \binits{S.L.}},
\oauthor{\bsnm{Eickhoff}, \binits{C.}}:
Neural document embeddings for intensive care patient mortality prediction.
arXiv preprint arXiv:1612.00467
(2016)
\end{botherref}
\endbibitem

\bibitem[\protect\citeauthoryear{Huang et~al.}{2020}]{huang2020fusion}
\begin{barticle}
\bauthor{\bsnm{Huang}, \binits{S.-C.}},
\bauthor{\bsnm{Pareek}, \binits{A.}},
\bauthor{\bsnm{Seyyedi}, \binits{S.}},
\bauthor{\bsnm{Banerjee}, \binits{I.}},
\bauthor{\bsnm{Lungren}, \binits{M.P.}}:
\batitle{Fusion of medical imaging and electronic health records using deep
  learning: a systematic review and implementation guidelines}.
\bjtitle{npj Digital Medicine}
\bvolume{3}(\bissue{1}),
\bfpage{136}
(\byear{2020})
\doiurl{10.1038/s41746-020-00341-z}
\end{barticle}
\endbibitem

\bibitem[\protect\citeauthoryear{Liu et~al.}{2017}]{liu2017visual}
\begin{barticle}
\bauthor{\bsnm{Liu}, \binits{N.}},
\bauthor{\bsnm{Wang}, \binits{K.}},
\bauthor{\bsnm{Jin}, \binits{X.}},
\bauthor{\bsnm{Gao}, \binits{B.}},
\bauthor{\bsnm{Dellandr{\'e}a}, \binits{E.}},
\bauthor{\bsnm{Chen}, \binits{L.}}:
\batitle{Visual affective classification by combining visual and text
  features}.
\bjtitle{PloS one}
\bvolume{12}(\bissue{8}),
\bfpage{0183018}
(\byear{2017})
\end{barticle}
\endbibitem

\bibitem[\protect\citeauthoryear{Liu et~al.}{2018a}]{liu2018prediction}
\begin{barticle}
\bauthor{\bsnm{Liu}, \binits{J.}},
\bauthor{\bsnm{Chen}, \binits{Y.}},
\bauthor{\bsnm{Lan}, \binits{L.}},
\bauthor{\bsnm{Lin}, \binits{B.}},
\bauthor{\bsnm{Chen}, \binits{W.}},
\bauthor{\bsnm{Wang}, \binits{M.}},
\bauthor{\bsnm{Li}, \binits{R.}},
\bauthor{\bsnm{Yang}, \binits{Y.}},
\bauthor{\bsnm{Zhao}, \binits{B.}},
\bauthor{\bsnm{Hu}, \binits{Z.}}, \betal:
\batitle{Prediction of rupture risk in anterior communicating artery aneurysms
  with a feed-forward artificial neural network}.
\bjtitle{European radiology}
\bvolume{28}(\bissue{8}),
\bfpage{3268}--\blpage{3275}
(\byear{2018})
\end{barticle}
\endbibitem

\bibitem[\protect\citeauthoryear{Liu et~al.}{2018b}]{liu2018bone}
\begin{barticle}
\bauthor{\bsnm{Liu}, \binits{M.}},
\bauthor{\bsnm{Lan}, \binits{J.}},
\bauthor{\bsnm{Chen}, \binits{X.}}, \betal:
\batitle{Bone age assessment model based on multi-dimensional feature fusion
  using deep learning}.
\bjtitle{Academic Journal of Second Military Medical University}
\bvolume{39},
\bfpage{909}--\blpage{916}
(\byear{2018})
\end{barticle}
\endbibitem

\bibitem[\protect\citeauthoryear{Bakkali et~al.}{2020}]{bakkali2020visual}
\begin{bchapter}
\bauthor{\bsnm{Bakkali}, \binits{S.}},
\bauthor{\bsnm{Ming}, \binits{Z.}},
\bauthor{\bsnm{Coustaty}, \binits{M.}},
\bauthor{\bsnm{Rusinol}, \binits{M.}}:
\bctitle{Visual and textual deep feature fusion for document image
  classification}.
In: \bbtitle{Proceedings of the IEEE/CVF Conference on Computer Vision and
  Pattern Recognition Workshops},
pp. \bfpage{562}--\blpage{563}
(\byear{2020})
\end{bchapter}
\endbibitem

\bibitem[\protect\citeauthoryear{Reda et~al.}{2018}]{reda2018deep}
\begin{barticle}
\bauthor{\bsnm{Reda}, \binits{I.}},
\bauthor{\bsnm{Khalil}, \binits{A.}},
\bauthor{\bsnm{Elmogy}, \binits{M.}},
\bauthor{\bsnm{Abou~El-Fetouh}, \binits{A.}},
\bauthor{\bsnm{Shalaby}, \binits{A.}},
\bauthor{\bsnm{Abou~El-Ghar}, \binits{M.}},
\bauthor{\bsnm{Elmaghraby}, \binits{A.}},
\bauthor{\bsnm{Ghazal}, \binits{M.}},
\bauthor{\bsnm{El-Baz}, \binits{A.}}:
\batitle{Deep learning role in early diagnosis of prostate cancer}.
\bjtitle{Technology in cancer research \& treatment}
\bvolume{17},
\bfpage{1533034618775530}
(\byear{2018})
\end{barticle}
\endbibitem

\bibitem[\protect\citeauthoryear{Qiu et~al.}{2018}]{qiu2018fusion}
\begin{barticle}
\bauthor{\bsnm{Qiu}, \binits{S.}},
\bauthor{\bsnm{Chang}, \binits{G.H.}},
\bauthor{\bsnm{Panagia}, \binits{M.}},
\bauthor{\bsnm{Gopal}, \binits{D.M.}},
\bauthor{\bsnm{Au}, \binits{R.}},
\bauthor{\bsnm{Kolachalama}, \binits{V.B.}}:
\batitle{Fusion of deep learning models of mri scans, mini--mental state
  examination, and logical memory test enhances diagnosis of mild cognitive
  impairment}.
\bjtitle{Alzheimer's \& Dementia: Diagnosis, Assessment \& Disease Monitoring}
\bvolume{10},
\bfpage{737}--\blpage{749}
(\byear{2018})
\end{barticle}
\endbibitem

\bibitem[\protect\citeauthoryear{Yala et~al.}{2019}]{yala2019deep}
\begin{barticle}
\bauthor{\bsnm{Yala}, \binits{A.}},
\bauthor{\bsnm{Lehman}, \binits{C.}},
\bauthor{\bsnm{Schuster}, \binits{T.}},
\bauthor{\bsnm{Portnoi}, \binits{T.}},
\bauthor{\bsnm{Barzilay}, \binits{R.}}:
\batitle{A deep learning mammography-based model for improved breast cancer
  risk prediction}.
\bjtitle{Radiology}
\bvolume{292}(\bissue{1}),
\bfpage{60}--\blpage{66}
(\byear{2019})
\end{barticle}
\endbibitem

\bibitem[\protect\citeauthoryear{Kawahara et~al.}{2018}]{kawahara2018seven}
\begin{barticle}
\bauthor{\bsnm{Kawahara}, \binits{J.}},
\bauthor{\bsnm{Daneshvar}, \binits{S.}},
\bauthor{\bsnm{Argenziano}, \binits{G.}},
\bauthor{\bsnm{Hamarneh}, \binits{G.}}:
\batitle{Seven-point checklist and skin lesion classification using multitask
  multimodal neural nets}.
\bjtitle{IEEE journal of biomedical and health informatics}
\bvolume{23}(\bissue{2}),
\bfpage{538}--\blpage{546}
(\byear{2018})
\end{barticle}
\endbibitem

\bibitem[\protect\citeauthoryear{Yoo et~al.}{2019}]{yoo2019deep}
\begin{barticle}
\bauthor{\bsnm{Yoo}, \binits{Y.}},
\bauthor{\bsnm{Tang}, \binits{L.Y.}},
\bauthor{\bsnm{Li}, \binits{D.K.}},
\bauthor{\bsnm{Metz}, \binits{L.}},
\bauthor{\bsnm{Kolind}, \binits{S.}},
\bauthor{\bsnm{Traboulsee}, \binits{A.L.}},
\bauthor{\bsnm{Tam}, \binits{R.C.}}:
\batitle{Deep learning of brain lesion patterns and user-defined clinical and
  mri features for predicting conversion to multiple sclerosis from clinically
  isolated syndrome}.
\bjtitle{Computer Methods in Biomechanics and Biomedical Engineering: Imaging
  \& Visualization}
\bvolume{7}(\bissue{3}),
\bfpage{250}--\blpage{259}
(\byear{2019})
\end{barticle}
\endbibitem

\bibitem[\protect\citeauthoryear{Ford et~al.}{2016}]{ford2016extracting}
\begin{barticle}
\bauthor{\bsnm{Ford}, \binits{E.}},
\bauthor{\bsnm{Carroll}, \binits{J.A.}},
\bauthor{\bsnm{Smith}, \binits{H.E.}},
\bauthor{\bsnm{Scott}, \binits{D.}},
\bauthor{\bsnm{Cassell}, \binits{J.A.}}:
\batitle{Extracting information from the text of electronic medical records to
  improve case detection: a systematic review}.
\bjtitle{Journal of the American Medical Informatics Association}
\bvolume{23}(\bissue{5}),
\bfpage{1007}--\blpage{1015}
(\byear{2016})
\end{barticle}
\endbibitem

\bibitem[\protect\citeauthoryear{Weissman et~al.}{2018}]{weissman2018inclusion}
\begin{barticle}
\bauthor{\bsnm{Weissman}, \binits{G.E.}},
\bauthor{\bsnm{Hubbard}, \binits{R.A.}},
\bauthor{\bsnm{Ungar}, \binits{L.H.}},
\bauthor{\bsnm{Harhay}, \binits{M.O.}},
\bauthor{\bsnm{Greene}, \binits{C.S.}},
\bauthor{\bsnm{Himes}, \binits{B.E.}},
\bauthor{\bsnm{Halpern}, \binits{S.D.}}:
\batitle{Inclusion of unstructured clinical text improves early prediction of
  death or prolonged icu stay}.
\bjtitle{Critical care medicine}
\bvolume{46}(\bissue{7}),
\bfpage{1125}
(\byear{2018})
\end{barticle}
\endbibitem

\bibitem[\protect\citeauthoryear{Johnson et~al.}{}]{johnsonalistairmimiciv}
\begin{botherref}
\oauthor{\bsnm{Johnson}, \binits{A.}},
\oauthor{\bsnm{Bulgarelli}, \binits{L.}},
\oauthor{\bsnm{Pollard}, \binits{T.}},
\oauthor{\bsnm{Horng}, \binits{S.}},
\oauthor{\bsnm{Celi}, \binits{L.A.}},
\oauthor{\bsnm{Mark}, \binits{R.}}:
{{MIMIC}}-{{IV}}.
{PhysioNet}.
\doiurl{10.13026/A3WN-HQ05}
\end{botherref}
\endbibitem

\bibitem[\protect\citeauthoryear{Wang et~al.}{2017}]{wang2017chestxray8}
\begin{bchapter}
\bauthor{\bsnm{Wang}, \binits{X.}},
\bauthor{\bsnm{Peng}, \binits{Y.}},
\bauthor{\bsnm{Lu}, \binits{L.}},
\bauthor{\bsnm{Lu}, \binits{Z.}},
\bauthor{\bsnm{Bagheri}, \binits{M.}},
\bauthor{\bsnm{Summers}, \binits{R.M.}}:
\bctitle{Chestx-ray8: hospital-scale chest x-ray database and benchmarks on
  weakly-supervised classification and localization of common thorax diseases}.
In: \bbtitle{{{IEEE Conference}} on {{Computer Vision}} and {{Pattern
  Recognition}} ({{CVPR}})},
pp. \bfpage{3462}--\blpage{3471}.
\bpublisher{{IEEE}}, \blocation{???}
(\byear{2017}).
\doiurl{10.1109/CVPR.2017.369}
\end{bchapter}
\endbibitem

\bibitem[\protect\citeauthoryear{Katzman et~al.}{2018}]{katzman2018deepsurv}
\begin{barticle}
\bauthor{\bsnm{Katzman}, \binits{J.L.}},
\bauthor{\bsnm{Shaham}, \binits{U.}},
\bauthor{\bsnm{Cloninger}, \binits{A.}},
\bauthor{\bsnm{Bates}, \binits{J.}},
\bauthor{\bsnm{Jiang}, \binits{T.}},
\bauthor{\bsnm{Kluger}, \binits{Y.}}:
\batitle{Deepsurv: personalized treatment recommender system using a cox
  proportional hazards deep neural network}.
\bjtitle{BMC medical research methodology}
\bvolume{18}(\bissue{1}),
\bfpage{1}--\blpage{12}
(\byear{2018})
\end{barticle}
\endbibitem

\bibitem[\protect\citeauthoryear{Irvin et~al.}{2019}]{irvin2019chexpert}
\begin{bchapter}
\bauthor{\bsnm{Irvin}, \binits{J.}},
\bauthor{\bsnm{Rajpurkar}, \binits{P.}},
\bauthor{\bsnm{Ko}, \binits{M.}},
\bauthor{\bsnm{Yu}, \binits{Y.}},
\bauthor{\bsnm{Ciurea-Ilcus}, \binits{S.}},
\bauthor{\bsnm{Chute}, \binits{C.}},
\bauthor{\bsnm{Marklund}, \binits{H.}},
\bauthor{\bsnm{Haghgoo}, \binits{B.}},
\bauthor{\bsnm{Ball}, \binits{R.}},
\bauthor{\bsnm{Shpanskaya}, \binits{K.}}, \betal:
\bctitle{Chexpert: A large chest radiograph dataset with uncertainty labels and
  expert comparison}.
In: \bbtitle{Proceedings of the AAAI Conference on Artificial Intelligence},
vol. \bseriesno{33},
pp. \bfpage{590}--\blpage{597}
(\byear{2019})
\end{bchapter}
\endbibitem

\bibitem[\protect\citeauthoryear{Johnson et~al.}{2019}]{johnson2019mimiccxrjpg}
\begin{botherref}
\oauthor{\bsnm{Johnson}, \binits{A.E.W.}},
\oauthor{\bsnm{Pollard}, \binits{T.J.}},
\oauthor{\bsnm{Greenbaum}, \binits{N.R.}},
\oauthor{\bsnm{Lungren}, \binits{M.P.}},
\oauthor{\bsnm{Deng}, \binits{C.-y.}},
\oauthor{\bsnm{Peng}, \binits{Y.}},
\oauthor{\bsnm{Lu}, \binits{Z.}},
\oauthor{\bsnm{Mark}, \binits{R.G.}},
\oauthor{\bsnm{Berkowitz}, \binits{S.J.}},
\oauthor{\bsnm{Horng}, \binits{S.}}:
{{MIMIC}}-{{CXR}}-{{JPG}}, a large publicly available database of labeled chest
  radiographs.
arXiv preprint
(2019)
{\href{https://arxiv.org/abs/1901.07042v5}{{arXiv:1901.07042v5}}}
\end{botherref}
\endbibitem

\bibitem[\protect\citeauthoryear{Peng et~al.}{2018}]{peng2018negbio}
\begin{bchapter}
\bauthor{\bsnm{Peng}, \binits{Y.}},
\bauthor{\bsnm{Wang}, \binits{X.}},
\bauthor{\bsnm{Lu}, \binits{L.}},
\bauthor{\bsnm{Bagheri}, \binits{M.}},
\bauthor{\bsnm{Summers}, \binits{R.}},
\bauthor{\bsnm{Lu}, \binits{Z.}}:
\bctitle{{{NegBio}}: a high-performance tool for negation and uncertainty
  detection in radiology reports.}
In: \bbtitle{{{AMIA Joint Summits}} on {{Translational Science}} Proceedings.
  {{AMIA Joint Summits}} on {{Translational Science}}},
vol. \bseriesno{2017},
pp. \bfpage{188}--\blpage{196}
(\byear{2018})
\end{bchapter}
\endbibitem

\bibitem[\protect\citeauthoryear{Zhang et~al.}{2020}]{Zhang2020WhenRR}
\begin{botherref}
\oauthor{\bsnm{Zhang}, \binits{Y.}},
\oauthor{\bsnm{Wang}, \binits{X.}},
\oauthor{\bsnm{Xu}, \binits{Z.}},
\oauthor{\bsnm{Yu}, \binits{Q.}},
\oauthor{\bsnm{Yuille}, \binits{A.}},
\oauthor{\bsnm{Xu}, \binits{D.}}:
When radiology report generation meets knowledge graph.
ArXiv
\textbf{abs/2002.08277}
(2020)
\end{botherref}
\endbibitem

\bibitem[\protect\citeauthoryear{Kipf and Welling}{2017}]{Kipf:2016tc}
\begin{bchapter}
\bauthor{\bsnm{Kipf}, \binits{T.N.}},
\bauthor{\bsnm{Welling}, \binits{M.}}:
\bctitle{{Semi-Supervised Classification with Graph Convolutional Networks}}.
In: \bbtitle{Proceedings of the 5th International Conference on Learning
  Representations}.
\bsertitle{ICLR '17}
(\byear{2017}).
\burl{https://openreview.net/forum?id=SJU4ayYgl}
\end{bchapter}
\endbibitem

\bibitem[\protect\citeauthoryear{Huang et~al.}{2017}]{huang2017densely}
\begin{bchapter}
\bauthor{\bsnm{Huang}, \binits{G.}},
\bauthor{\bsnm{Liu}, \binits{Z.}},
\bauthor{\bsnm{Van Der~Maaten}, \binits{L.}},
\bauthor{\bsnm{Weinberger}, \binits{K.Q.}}:
\bctitle{Densely connected convolutional networks}.
In: \bbtitle{Proceedings of the IEEE Conference on Computer Vision and Pattern
  Recognition},
pp. \bfpage{4700}--\blpage{4708}
(\byear{2017})
\end{bchapter}
\endbibitem

\bibitem[\protect\citeauthoryear{Rajpurkar et~al.}{2017}]{rajpurkar2017chexnet}
\begin{botherref}
\oauthor{\bsnm{Rajpurkar}, \binits{P.}},
\oauthor{\bsnm{Irvin}, \binits{J.}},
\oauthor{\bsnm{Zhu}, \binits{K.}},
\oauthor{\bsnm{Yang}, \binits{B.}},
\oauthor{\bsnm{Mehta}, \binits{H.}},
\oauthor{\bsnm{Duan}, \binits{T.}},
\oauthor{\bsnm{Ding}, \binits{D.}},
\oauthor{\bsnm{Bagul}, \binits{A.}},
\oauthor{\bsnm{Langlotz}, \binits{C.}},
\oauthor{\bsnm{Shpanskaya}, \binits{K.}}, et al.:
Chexnet: Radiologist-level pneumonia detection on chest x-rays with deep
  learning.
arXiv preprint arXiv:1711.05225
(2017)
\end{botherref}
\endbibitem

\bibitem[\protect\citeauthoryear{Kingma and Ba}{2014}]{kingma2014adam}
\begin{botherref}
\oauthor{\bsnm{Kingma}, \binits{D.P.}},
\oauthor{\bsnm{Ba}, \binits{J.}}:
Adam: A method for stochastic optimization.
arXiv preprint arXiv:1412.6980
(2014)
\end{botherref}
\endbibitem

\bibitem[\protect\citeauthoryear{Peng et~al.}{2019}]{peng2019transfer}
\begin{bchapter}
\bauthor{\bsnm{Peng}, \binits{Y.}},
\bauthor{\bsnm{Yan}, \binits{S.}},
\bauthor{\bsnm{Lu}, \binits{Z.}}:
\bctitle{Transfer learning in biomedical natural language processing: An
  evaluation of bert and elmo on ten benchmarking datasets}.
In: \bbtitle{Proceedings of the 2019 Workshop on Biomedical Natural Language
  Processing (BioNLP 2019)},
pp. \bfpage{58}--\blpage{65}
(\byear{2019})
\end{bchapter}
\endbibitem

\bibitem[\protect\citeauthoryear{P{\"o}lsterl}{2020}]{polsterl2020scikit}
\begin{barticle}
\bauthor{\bsnm{P{\"o}lsterl}, \binits{S.}}:
\batitle{scikit-survival: A library for time-to-event analysis built on top of
  scikit-learn}.
\bjtitle{The Journal of Machine Learning Research}
\bvolume{21}(\bissue{1}),
\bfpage{8747}--\blpage{8752}
(\byear{2020})
\end{barticle}
\endbibitem

\bibitem[\protect\citeauthoryear{Cortes et~al.}{2008}]{cortes2008sample}
\begin{bchapter}
\bauthor{\bsnm{Cortes}, \binits{C.}},
\bauthor{\bsnm{Mohri}, \binits{M.}},
\bauthor{\bsnm{Riley}, \binits{M.}},
\bauthor{\bsnm{Rostamizadeh}, \binits{A.}}:
\bctitle{Sample selection bias correction theory}.
In: \bbtitle{Algorithmic Learning Theory: 19th International Conference, ALT
  2008, Budapest, Hungary, October 13-16, 2008. Proceedings 19},
pp. \bfpage{38}--\blpage{53}
(\byear{2008}).
\bcomment{Springer}
\end{bchapter}
\endbibitem

\bibitem[\protect\citeauthoryear{Whittemore}{1978}]{whittemore1978collapsibility}
\begin{barticle}
\bauthor{\bsnm{Whittemore}, \binits{A.S.}}:
\batitle{Collapsibility of multidimensional contingency tables}.
\bjtitle{Journal of the Royal Statistical Society: Series B (Methodological)}
\bvolume{40}(\bissue{3}),
\bfpage{328}--\blpage{340}
(\byear{1978})
\end{barticle}
\endbibitem

\bibitem[\protect\citeauthoryear{Robins}{2001}]{robins2001data}
\begin{botherref}
\oauthor{\bsnm{Robins}, \binits{J.M.}}:
Data, design, and background knowledge in etiologic inference.
Epidemiology,
313--320
(2001)
\end{botherref}
\endbibitem

\bibitem[\protect\citeauthoryear{Goodfellow
  et~al.}{2014}]{goodfellow2014explaining}
\begin{botherref}
\oauthor{\bsnm{Goodfellow}, \binits{I.J.}},
\oauthor{\bsnm{Shlens}, \binits{J.}},
\oauthor{\bsnm{Szegedy}, \binits{C.}}:
Explaining and harnessing adversarial examples.
arXiv preprint arXiv:1412.6572
(2014)
\end{botherref}
\endbibitem

\bibitem[\protect\citeauthoryear{Paschali
  et~al.}{2018}]{paschali2018generalizability}
\begin{bchapter}
\bauthor{\bsnm{Paschali}, \binits{M.}},
\bauthor{\bsnm{Conjeti}, \binits{S.}},
\bauthor{\bsnm{Navarro}, \binits{F.}},
\bauthor{\bsnm{Navab}, \binits{N.}}:
\bctitle{Generalizability vs. robustness: investigating medical imaging
  networks using adversarial examples}.
In: \bbtitle{Medical Image Computing and Computer Assisted Intervention--MICCAI
  2018: 21st International Conference, Granada, Spain, September 16-20, 2018,
  Proceedings, Part I},
pp. \bfpage{493}--\blpage{501}
(\byear{2018}).
\bcomment{Springer}
\end{bchapter}
\endbibitem

\bibitem[\protect\citeauthoryear{Finlayson
  et~al.}{2019}]{finlayson2019adversarial}
\begin{barticle}
\bauthor{\bsnm{Finlayson}, \binits{S.G.}},
\bauthor{\bsnm{Bowers}, \binits{J.D.}},
\bauthor{\bsnm{Ito}, \binits{J.}},
\bauthor{\bsnm{Zittrain}, \binits{J.L.}},
\bauthor{\bsnm{Beam}, \binits{A.L.}},
\bauthor{\bsnm{Kohane}, \binits{I.S.}}:
\batitle{Adversarial attacks on medical machine learning}.
\bjtitle{Science}
\bvolume{363}(\bissue{6433}),
\bfpage{1287}--\blpage{1289}
(\byear{2019})
\end{barticle}
\endbibitem

\end{thebibliography}

\end{document}